\documentclass[11pt]{article}

\usepackage[preprint]{acl}

\usepackage{times}
\usepackage{latexsym}

\usepackage[T1]{fontenc}
\usepackage[utf8]{inputenc}

\usepackage{microtype}

\usepackage{inconsolata}

\usepackage{graphicx}

\usepackage{amsfonts}
\usepackage{amsmath}
\usepackage{amssymb}

\usepackage{booktabs}
\usepackage{multirow} 
\title{Stepwise Reasoning Enhancement for LLMs via External Subgraph Generation}

\author{Xin Zhang\textsuperscript{1}, Yang Cao\textsuperscript{2}, Baoxing Wu\textsuperscript{1}, Kai Song\textsuperscript{2}, Siying Li\textsuperscript{1} \\
$^1$School of Information Science and Engineering, Chongqing Jiaotong University\\ $^2$School of Computer Science and Technology, Chongqing University of Posts and Telecommunications\\
}

\begin{document}
\maketitle
\begin{abstract}
Large language models have shown strong performance in natural language generation and downstream reasoning tasks, but they still struggle with logical consistency, factual grounding, and interpretability in complex multi-step reasoning. To address these limitations, this paper proposes SGR, a stepwise reasoning enhancement framework that integrates large language models with external knowledge graphs through query-relevant subgraph generation. Given an input question, SGR first extracts key entities, relations, and constraints to construct a structured schema, then retrieves compact subgraphs from a knowledge graph using schema-guided querying. The generated subgraphs provide explicit relational evidence that guides the language model through step-by-step reasoning. In addition, SGR combines direct Cypher-based reasoning with collaborative reasoning integration, allowing candidate answers from multiple reasoning paths to be validated and aggregated according to both model confidence and graph consistency. Experiments on benchmark datasets including CWQ, WebQSP, GrailQA, and KQA Pro demonstrate that SGR improves reasoning accuracy and Hits@1 performance over standard prompting and several knowledge-enhanced baselines. Ablation studies further show that schema guidance and Neo4j-based retrieval are both crucial to the effectiveness of the framework. These results indicate that dynamically generated external subgraphs can improve the accuracy, robustness, and interpretability of LLM-based reasoning.
\end{abstract}

\section{Introduction}

Large language models (LLMs) have achieved remarkable success in natural language understanding, generation, and few-shot learning, demonstrating strong generalization across a wide range of downstream tasks \citep{brown2020language}. Recent prompting strategies, such as chain-of-thought reasoning, further enable LLMs to decompose complex problems into intermediate steps and improve their performance on multi-step reasoning tasks \citep{wei2022chain}. Despite these advances, LLMs still face important limitations when solving knowledge-intensive reasoning problems. Their outputs may lack factual grounding, suffer from logical inconsistency, or provide reasoning processes that are difficult to verify. These issues are especially problematic in complex question answering scenarios where correct answers depend on multiple entities, relations, and constraints. Moreover, because the reasoning process of LLMs is often implicit, it remains challenging to ensure faithful interpretability and trace the evidence supporting a generated answer \citep{jacovi2020towards}.

Knowledge graphs provide a promising way to address these challenges by representing entities and relations as structured triples, enabling explicit and verifiable reasoning over external knowledge \citep{ji2022survey}. In knowledge-intensive question answering, structured resources such as Freebase have been widely used to support multi-hop inference and improve factual reliability \citep{yao2014information}. Retrieval-augmented generation further demonstrates that supplying external evidence to language models can improve their performance on knowledge-intensive tasks \citep{lewis2020retrieval}. However, conventional retrieval methods often rely on unstructured passages or loosely connected facts, which may fail to preserve the relational dependencies required for complex reasoning. Recent studies have therefore explored more structured forms of reasoning, including logic-guided translation, deliberate planning, and graph-based interaction between LLMs and knowledge sources \citep{yang2024harnessing,xiong2025deliberate,jiang2023structgpt,li2023tog}. Nevertheless, how to dynamically construct compact, query-relevant graph evidence and use it to guide faithful stepwise reasoning remains an open problem.

To address this problem, we propose SGR, a stepwise reasoning enhancement framework that improves LLM reasoning by dynamically generating external subgraphs from knowledge graphs. Instead of relying only on the implicit knowledge stored in model parameters or retrieving isolated textual evidence, SGR first converts an input question into a structured schema containing key entities, relations, and constraints. This schema is then used to retrieve a compact query-relevant subgraph from an external knowledge graph. The generated subgraph provides explicit relational evidence, allowing the LLM to perform reasoning step by step along verifiable knowledge paths.

SGR further combines two complementary reasoning strategies. First, it performs direct reasoning enhancement by translating the generated schema into Cypher queries and executing them in Neo4j to obtain candidate answers grounded in the knowledge graph. Second, it applies collaborative reasoning integration, in which candidate reasoning paths are validated and aggregated according to both language model confidence and graph consistency. In this way, SGR not only improves factual grounding, but also reduces the risk of unsupported reasoning and enhances the interpretability of the final answer.

The main contributions of this paper are summarized as follows. First, we propose a stepwise reasoning enhancement framework that integrates LLMs with dynamically generated external subgraphs for knowledge-intensive question answering. Second, we introduce a schema-guided subgraph generation strategy that extracts query-relevant entities, relations, and constraints to retrieve compact structured evidence from knowledge graphs. Third, we design a collaborative reasoning integration mechanism that combines Cypher-based direct reasoning with graph-consistency validation across multiple reasoning paths. Finally, experiments on CWQ, WebQSP, GrailQA, and KQA Pro demonstrate that SGR improves Hits@1 and accuracy over standard prompting methods and several knowledge-enhanced baselines, while ablation studies confirm the importance of schema guidance and Neo4j-based retrieval.

\section{Related Work}

\subsection{Knowledge-Enhanced Language Models}

Knowledge graphs have long been used as structured resources for representing entities, relations, and factual triples. Early studies on relational machine learning over knowledge graphs provide the foundation for embedding entities and relations into continuous spaces and performing link prediction and reasoning over structured knowledge \citep{nickel2016review}. With the development of pre-trained language models, researchers have explored how to inject external knowledge into language representations. ERNIE incorporates knowledge information into BERT-style pre-training, showing that entity-aware representation learning can improve language understanding tasks \citep{sun2019bert}. More broadly, knowledge-enhanced pre-trained language models aim to combine symbolic knowledge with distributed representations, enabling models to better capture factual and relational information beyond text-only pre-training \citep{liu2023survey}.

In addition to representation learning, logical reasoning over knowledge graphs has been investigated as a way to improve interpretability and multi-hop inference. TILP learns temporal logical rules over knowledge graphs in a differentiable manner, demonstrating the value of explicit rule structures for temporal reasoning \citep{xiongtilp}. TEILP further extends this direction by using logical reasoning for time prediction over knowledge graphs \citep{xiong2024teilp}. Recent work also shows that large language models can acquire temporal reasoning ability when properly trained or prompted, suggesting that neural models and structured temporal knowledge can complement each other \citep{xiong2024large}. These studies motivate our use of external subgraphs as explicit structured evidence for improving the factual grounding and interpretability of LLM reasoning.

\subsection{Retrieval-Augmented and Knowledge-Intensive Reasoning}

Retrieval-augmented models improve language generation by providing external evidence at inference or pre-training time. Retrieval-Augmented Generation retrieves relevant passages from an external corpus and conditions generation on the retrieved evidence, improving performance on knowledge-intensive NLP tasks \citep{lewis2020retrieval}. REALM introduces retrieval into language model pre-training, allowing the model to retrieve documents from a large corpus and use them for prediction \citep{guu2020retrieval}. In open-domain question answering, multi-step retriever-reader frameworks iteratively retrieve and read evidence, which helps answer complex questions requiring multiple supporting facts \citep{das2019multi}. Fusion-in-Decoder further shows that generative models can effectively leverage multiple retrieved passages by encoding them separately and fusing the evidence during decoding \citep{izacard2021leveraging}.

Several benchmarks and systems have been proposed to evaluate knowledge-intensive reasoning. KILT provides a unified benchmark for knowledge-intensive language tasks and emphasizes the importance of grounding model outputs in external sources \citep{petroni2021kilt}. Open question answering over tables and text explores reasoning across heterogeneous evidence types, showing that complex QA often requires combining structured and unstructured information \citep{chen2020open}. Knowledge-augmented prompting methods further demonstrate that LLMs can benefit from explicit knowledge retrieved from knowledge graphs in zero-shot KGQA settings \citep{logan2019barack}. Compared with these approaches, our framework focuses on dynamically generating compact query-relevant subgraphs rather than retrieving isolated text passages or loosely connected facts. This allows the model to reason over explicit relational paths and better preserve the structural dependencies required for multi-hop question answering.

\subsection{Structured and Collaborative Reasoning with LLMs}

Recent work has increasingly explored how to guide LLMs with structured reasoning processes. Least-to-most prompting decomposes complex problems into simpler subproblems and solves them sequentially, showing that stepwise reasoning can improve model performance on difficult tasks \citep{zhou2022least}. StructGPT provides a general framework for enabling LLMs to reason over structured data through iterative reading and reasoning operations \citep{jiang2023structgpt}. Graph reasoning enhanced language models further incorporate graph neural reasoning into language models for question answering, demonstrating the benefit of explicitly modeling relational structures \citep{luo2022knowledge}.

Another important direction is collaborative reasoning between LLMs and knowledge graphs. Recent surveys and roadmaps argue that LLMs and KGs have complementary strengths: LLMs provide flexible language understanding and generation, while KGs provide precise, interpretable, and verifiable structured knowledge \citep{yang2023collaborative}. This complementarity is especially important for complex knowledge-intensive QA, where reasoning often requires both semantic understanding of the question and faithful traversal of entity-relation paths. Our proposed SGR framework follows this direction by integrating LLM-based schema construction, Cypher-based graph querying, and collaborative answer validation. Unlike methods that rely only on retrieved passages or direct prompting, SGR uses generated external subgraphs to support stepwise reasoning and aggregates candidate answers according to both model confidence and graph consistency. In this way, SGR aims to improve not only answer accuracy, but also the transparency and reliability of the reasoning process.

\begin{figure*}[t]
  \centering
  \includegraphics[width=0.8\linewidth]{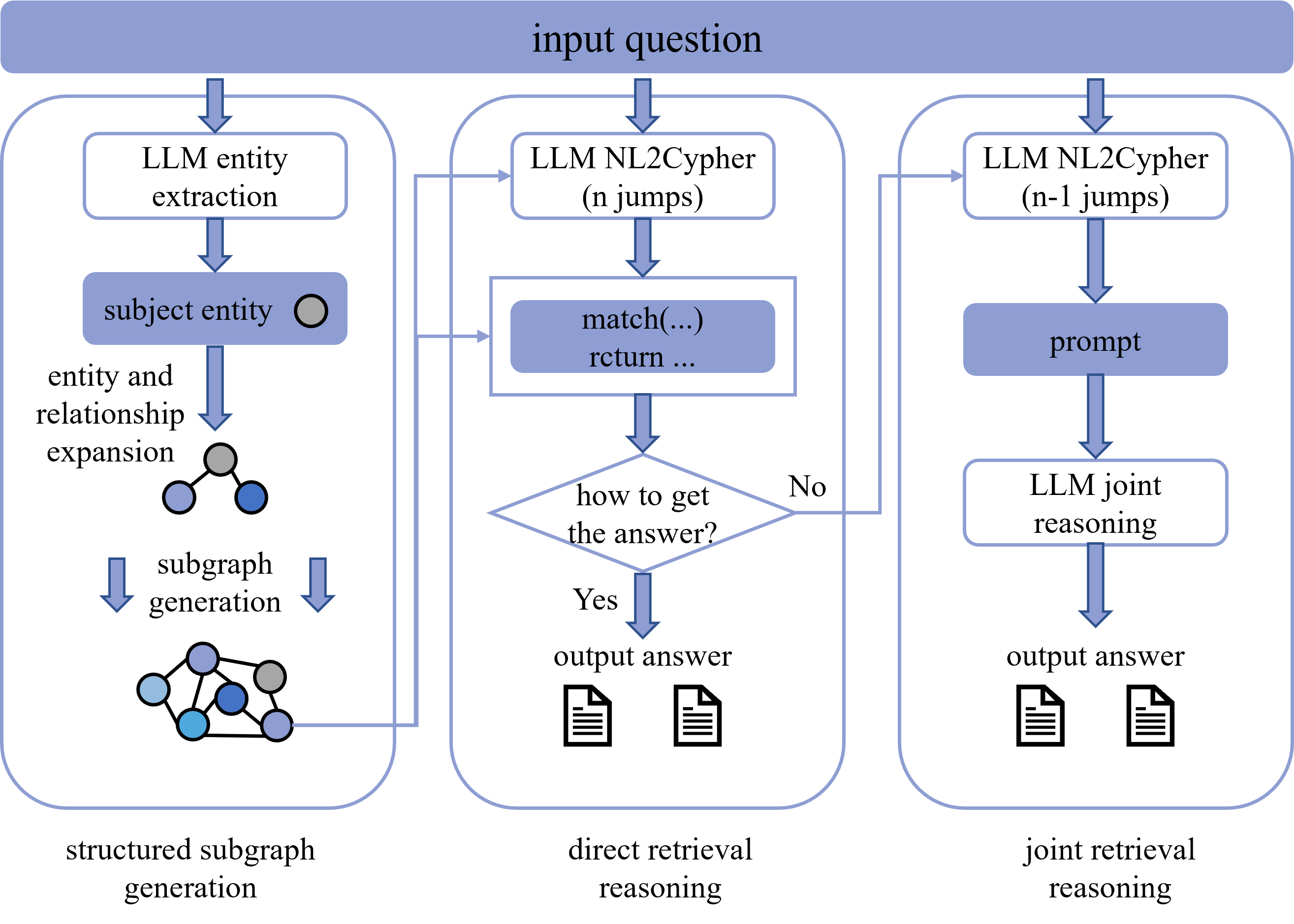}
  \caption{Pipeline of SGR framework.}
  \label{fig:fig1}
\end{figure*}

\section{Methodology}

\subsection{Structured Subgraph Generation}

\subsubsection{Knowledge Graph Construction}

SGR uses an external knowledge graph as the structured reasoning source for generating query-relevant evidence. The knowledge graph is represented as
\begin{equation}
\mathcal{G}=(\mathcal{V},\mathcal{E},\mathcal{R}),
\end{equation}
where $\mathcal{V}$ denotes the set of entities, $\mathcal{R}$ denotes the set of relation types, and $\mathcal{E}$ denotes the set of factual triples. Each edge in the graph is represented as a triple
\begin{equation}
e=(h,r,t),
\end{equation}
where $h,t \in \mathcal{V}$ are the head and tail entities, and $r \in \mathcal{R}$ is the relation connecting them. This triple-based representation allows SGR to explicitly model factual dependencies among entities and provides a verifiable basis for multi-hop reasoning.

To support efficient graph querying, the extracted triples are stored in Neo4j. Entities are modeled as graph nodes, while relations are modeled as directed edges between nodes. Each node contains attributes such as entity name, type, and identifier, and each edge contains the corresponding relation label and optional constraint information. This construction enables the system to retrieve paths, neighborhoods, and constrained relation patterns through Cypher queries.

Given an input question $q$, SGR first identifies the question-specific schema
\begin{equation}
\mathcal{S}_q = {\mathcal{V}_q,\mathcal{R}_q,\mathcal{C}_q},
\end{equation}
where $\mathcal{V}_q$ is the set of linked entities, $\mathcal{R}_q$ is the set of candidate relations, and $\mathcal{C}_q$ is the set of constraints, such as entity types, temporal restrictions, comparison conditions, or numerical limits. The schema serves as an intermediate structured representation between the natural language question and the external knowledge graph.

The entity linking module maps mentions in the question to corresponding nodes in $\mathcal{G}$. Relation extraction then identifies possible predicates that connect the linked entities to potential answer nodes. Constraint extraction further captures semantic conditions that must be satisfied during graph retrieval. For example, in questions involving time, location, ranking, or aggregation, the corresponding constraints are added to $\mathcal{C}_q$ and later used to filter candidate reasoning paths.

Through this construction process, the full knowledge graph is transformed into a searchable structured space. Instead of allowing the language model to reason only from implicit parametric knowledge, SGR grounds the reasoning process in explicit entities, relations, and constraints. This provides the foundation for the subsequent subgraph generation stage, where compact evidence subgraphs are dynamically retrieved according to the question schema.

\subsubsection{Subgraph Generation Process}

After constructing the question schema $\mathcal{S}_q$, SGR generates a compact query-relevant subgraph from the external knowledge graph. Starting from the linked entities in $\mathcal{V}_q$, the framework expands along candidate relations in $\mathcal{R}_q$ while applying the constraints in $\mathcal{C}_q$. This process retrieves entity-relation paths that are most relevant to the input question.

A candidate reasoning path is denoted as
\begin{equation}
p_i=(v_0,r_1,v_1,\ldots,r_k,v_k),
\end{equation}
where $v_0$ is a seed entity and $k$ is the path length. Each path is scored according to relation relevance, entity alignment, and constraint satisfaction. The top-ranked paths are merged to form the subgraph $\mathcal{G}_q$, while duplicate and low-relevance triples are removed. The resulting subgraph provides compact structured evidence for later reasoning and reduces noise from the full knowledge graph.

\subsection{Stepwise Reasoning Enhancement}

Given the generated subgraph $\mathcal{G}_q$, SGR guides the language model to reason step by step over explicit graph evidence. Instead of directly producing an answer, the model follows relevant entities, relations, and constraints in the subgraph to construct an interpretable reasoning trajectory.

Let $z_t$ denote the reasoning state at step $t$. The stepwise reasoning process is formulated as
\begin{equation}
z_t=f_{\theta}(q,\mathcal{S}_q,\mathcal{G}_q,z_{<t}),
\end{equation}
where $z_{<t}$ represents previous reasoning states. At each step, the model selects evidence from the subgraph and updates the reasoning state. Unsupported reasoning steps are filtered or assigned lower reliability. In this way, SGR improves factual grounding and makes the reasoning process easier to verify.

\subsection{Collaborative Reasoning Integration}

Complex questions may produce multiple candidate reasoning paths. To improve robustness, SGR integrates answers from different paths using both model confidence and graph consistency. For each path $p_i$, the language model produces a candidate answer $a_i$ with confidence $\rho_i$, while the graph module computes a consistency score $\eta_i$.

The reliability score of each path is defined as
\begin{equation}
\omega_i=\lambda \rho_i+(1-\lambda)\eta_i,
\end{equation}
where $\lambda$ balances the two signals. The final answer is selected by aggregating support from all paths:
\begin{equation}
\hat{a}=\arg\max_a \sum_i \omega_i \cdot \mathbb{I}(a_i=a).
\end{equation}

This mechanism suppresses noisy single-path predictions and favors answers supported by multiple reliable reasoning paths. Therefore, SGR improves both answer accuracy and interpretability.

\begin{table*}[t]
\centering
\small
\caption{Performance comparison of different reasoning methods on CWQ, WebQSP, and GrailQA. 
Hits@1 and accuracy are reported where applicable, and the best results for each metric are highlighted in bold.
Note: Best results are taken from prior work, including $\alpha$,
$\beta$,
$\gamma$,
and $\delta$.}
\label{tab:experimental_results}
\begin{tabular}{lcccccc}
\toprule
\multirow{2}{*}{Method} &
\multicolumn{2}{c}{CWQ} &
\multicolumn{2}{c}{WebQSP} &
\multicolumn{2}{c}{GrailQA} \\
\cmidrule(lr){2-3} \cmidrule(lr){4-5} \cmidrule(lr){6-7}
& Hits@1 & Acc & Hits@1 & Acc & Hits@1 & Acc \\
\midrule
IO Prompt / ChatGPT        & 0.376 & 0.256 & 0.633 & 0.582 & 0.294 & 0.223 \\
CoT / ChatGPT              & 0.388 & 0.258 & 0.622 & 0.577 & 0.281 & 0.201 \\
\midrule
Prior FT SOTA              & 0.704$^{\alpha}$ & -- & 0.821$^{\beta}$ & -- & 0.754$^{\gamma}$ & -- \\
Prior Prompting SOTA       & -- & -- & 0.744$^{\delta}$ & -- & 0.532$^{\delta}$ & -- \\
\midrule
SGR / Cypher LLM         & 0.523 & 0.445 & 0.745 & 0.706 & 0.624 & 0.593 \\
SGR / ChatGPT            & 0.578 & 0.526 & 0.801 & 0.784 & 0.713 & 0.633 \\
StructGPT / ChatGPT        & -- & -- & 0.726 & -- & -- & -- \\
ToG / ChatGPT              & 0.571 & -- & 0.762 & -- & 0.687 & -- \\
ToG / GPT-4                & \textbf{0.725} & -- & 0.826 & -- & \textbf{0.814} & -- \\
SGR / GPT-4              & 0.632 & \textbf{0.590} & \textbf{0.826} & \textbf{0.808} & 0.756 & \textbf{0.703} \\
\bottomrule
\end{tabular}
\end{table*}

\subsection{Direct Reasoning Enhancement}

Direct reasoning enhancement aims to obtain graph-grounded candidate answers by converting the generated schema into executable Cypher queries. Unlike pure prompting methods, which rely mainly on the implicit knowledge of the language model, this component performs explicit reasoning over the external knowledge graph. Given the structured schema $\mathcal{S}_q={\mathcal{V}_q,\mathcal{R}_q,\mathcal{C}_q}$, SGR constructs Cypher queries that search for entity-relation paths satisfying the semantic and logical constraints of the input question. The retrieved results provide direct evidence from Neo4j and serve as reliable candidate answers for subsequent validation and integration.

\subsubsection{Cypher LLM}

The Cypher LLM module translates the schema generated from the input question into a structured graph query. Specifically, the model receives the extracted entities, candidate relations, and constraints, and then generates a Cypher query that can be executed in Neo4j. This query is designed to match reasoning paths in the knowledge graph that are consistent with the question semantics.

Let the Cypher generation process be denoted as
\begin{equation}
\mathcal{Q}_c = g_{\theta}(q,\mathcal{S}_q),
\end{equation}
where $g_{\theta}$ represents the language model used for query generation and $\mathcal{Q}_c$ is the generated Cypher query. Executing $\mathcal{Q}_c$ on the external knowledge graph produces a set of candidate answers:
\begin{equation}
\mathcal{A}_c = \mathrm{Exec}(\mathcal{Q}_c,\mathcal{G}),
\end{equation}
where $\mathrm{Exec}(\cdot)$ denotes query execution in Neo4j.

The Cypher LLM module improves reasoning in two ways. First, it transforms natural language reasoning into executable symbolic operations, making the reasoning process more explicit and verifiable. Second, it narrows the search space by using the schema constraints, which helps avoid irrelevant entities and noisy relations. As a result, the model can obtain candidate answers that are directly grounded in structured graph evidence rather than relying only on generated textual reasoning.

\subsubsection{Answer Validation}

Although Cypher-based retrieval provides graph-grounded candidate answers, errors may still occur due to imperfect entity linking, relation matching, or query generation. Therefore, SGR introduces an answer validation step to verify whether each candidate answer is supported by the generated subgraph and consistent with the original question.

For each candidate answer $a_j \in \mathcal{A}_c$, SGR evaluates its validity using both graph evidence and semantic consistency. The graph consistency score measures whether the answer can be reached through a valid reasoning path in the generated subgraph:
\begin{equation}
\gamma_j = C(p_j,\mathcal{G}_q),
\end{equation}
where $p_j$ is the reasoning path associated with answer $a_j$. The semantic confidence score measures the compatibility between the question, the schema, and the candidate answer:
\begin{equation}
\delta_j = P_{\theta}(a_j \mid q,\mathcal{S}_q,p_j).
\end{equation}

The final validation score is computed as
\begin{equation}
v_j = \mu \delta_j + (1-\mu)\gamma_j,
\end{equation}
where $\mu \in [0,1]$ controls the balance between semantic confidence and graph consistency. Candidate answers with validation scores below a predefined threshold are filtered out:
\begin{equation}
\mathcal{A}_v = {a_j \in \mathcal{A}_c \mid v_j \geq \epsilon }.
\end{equation}

This validation mechanism reduces the influence of incorrect or weakly supported answers. By requiring candidate answers to be both semantically plausible and structurally supported by the knowledge graph, SGR improves the reliability of direct reasoning and provides more trustworthy inputs for the later collaborative reasoning stage.

\subsection{Collaborative Reasoning Enhancement}

To improve robustness, SGR integrates candidate answers from multiple reasoning paths rather than relying on a single Cypher query result. Given the query-relevant subgraph $\mathcal{G}_q$, the framework obtains a set of candidate reasoning paths:
\begin{equation}
\mathcal{P}_q = {p_1, p_2, \ldots, p_M}.
\end{equation}
Each path $p_i$ produces a candidate answer $a_i$ with a model confidence score:
\begin{equation}
\rho_i = P_{\theta}(a_i \mid q, p_i, \mathcal{S}_q),
\end{equation}
where $\mathcal{S}_q$ is the generated schema. SGR also evaluates the graph consistency of each path:
\begin{equation}
\eta_i = C(p_i, \mathcal{G}_q).
\end{equation}

The reliability score of each path is computed by combining model confidence and graph consistency:
\begin{equation}
\omega_i = \lambda \rho_i + (1-\lambda)\eta_i,
\end{equation}
where $\lambda \in [0,1]$ controls their relative importance. The final answer is selected by aggregating support from all paths:
\begin{equation}
\hat{a}=\arg\max_{a}
\sum_{i=1}^{M}
\omega_i \cdot \mathbb{I}(a_i = a).
\end{equation}

This collaborative reasoning strategy reduces the influence of noisy or incomplete paths. Answers supported by multiple high-confidence and graph-consistent paths receive higher scores, while unsupported answers are suppressed. As a result, SGR improves both the reliability and interpretability of multi-hop reasoning.

\section{Experiments}

\subsection{Experimental Setup}

We evaluate SGR on four knowledge-intensive question answering benchmarks: CWQ, WebQSP, GrailQA, and KQA Pro. These datasets test multi-hop, compositional, and structured reasoning over knowledge graphs. For each question, SGR extracts key entities, relations, and constraints to build a schema, which is then converted into a Cypher query and executed in Neo4j to retrieve relevant subgraphs and candidate answers.
We test three variants: SGR/Cypher LLM, SGR/ChatGPT, and SGR/GPT-4. We compare them with standard prompting methods, including IO prompting and Chain-of-Thought prompting, as well as knowledge-enhanced baselines such as StructGPT and Tree-of-Graphs. Performance is measured using Hits@1 and accuracy. We also conduct ablation studies by removing schema prompts and Neo4j retrieval to evaluate the contribution of each component.

\begin{table*}[t]
\centering
\small
\caption{Ablation experiment results on the CWQ dataset}
\label{tab:ablation_cwq}
\begin{tabular}{lcccccccc}
\toprule
\multirow{2}{*}{Method} &
\multicolumn{2}{c}{With Schema} &
\multicolumn{2}{c}{With neo4j} &
\multicolumn{2}{c}{Without Schema} &
\multicolumn{2}{c}{Without neo4j} \\
\cmidrule(lr){2-3} \cmidrule(lr){4-5} \cmidrule(lr){6-7} \cmidrule(lr){8-9}
 & Hits@1 & Acc & Hits@1 & Acc & Hits@1 & Acc & Hits@1 & Acc \\
\midrule
SGR/Cypher LLM & 0.523 & 0.445 & 0.553 & 0.445 & 0.322 & 0.128 & 0.360 & 0.238 \\
SGR/ChatGPT   & 0.578 & 0.526 & 0.578 & 0.526 & 0.400 & 0.319 & 0.431 & 0.374 \\
\bottomrule
\end{tabular}
\end{table*}

\subsection{Experimental Results}

Table~\ref{tab:experimental_results} presents the main results on CWQ, WebQSP, and GrailQA. Overall, SGR consistently outperforms standard prompting baselines, showing the effectiveness of using external subgraph evidence for multi-step reasoning.
On CWQ, SGR/ChatGPT improves Hits@1 from 0.376 for IO Prompt and 0.388 for CoT to 0.578, while accuracy increases to 0.526. With GPT-4, SGR further achieves the best accuracy of 0.590. On WebQSP, SGR/GPT-4 obtains 0.826 Hits@1 and 0.808 accuracy, matching the best Hits@1 result and achieving the highest accuracy among all methods. On GrailQA, SGR/GPT-4 reaches 0.756 Hits@1 and the best accuracy of 0.703, demonstrating strong generalization to complex query structures.

These results indicate that schema-guided subgraph generation and graph-grounded reasoning help LLMs produce more accurate and reliable answers. The consistent gains over prompting baselines and competitive performance against existing knowledge-enhanced methods confirm the effectiveness of the proposed SGR framework.

\subsection{Ablation Study}

The ablation study evaluates the effects of schema guidance and Neo4j-based retrieval on the CWQ dataset. As shown in Table~\ref{tab:ablation_cwq}, removing either component causes a clear performance drop. Without schema prompts, SGR/Cypher LLM decreases from 0.523 to 0.322 in Hits@1 and from 0.445 to 0.128 in accuracy, while SGR/ChatGPT drops from 0.578 to 0.400 in Hits@1 and from 0.526 to 0.319 in accuracy. This indicates that schema guidance is important for constructing accurate reasoning paths.
Similarly, removing Neo4j retrieval reduces performance for both models, showing that external graph evidence is necessary for factual grounding. Overall, the results demonstrate that schema prompts and Neo4j retrieval are complementary components, and their combination enables SGR to achieve more accurate and reliable multi-hop reasoning.

\begin{figure}[t]
  \centering
  \includegraphics[width=0.9\linewidth]{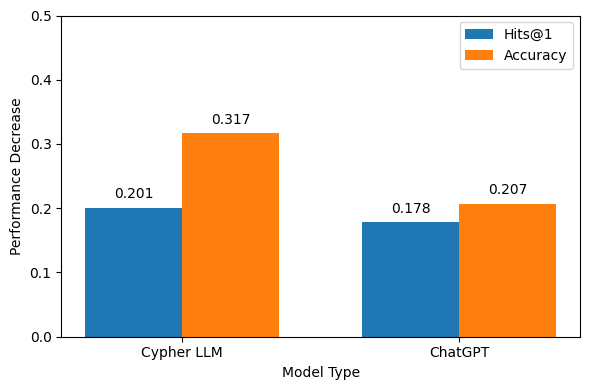}
  \vspace{-10pt}
  \caption{Impact brought by removing Schema prompts.}
  \includegraphics[width=0.9\linewidth]{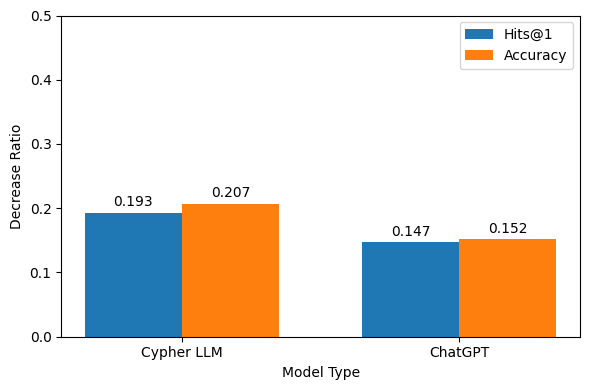}
  \vspace{-10pt}
  \caption{Impact brought by removing neo4j retrieval.}
\end{figure}

\subsection{Application Scenarios and Error Analysis}

SGR is suitable for knowledge-intensive tasks that require reliable multi-step reasoning, such as knowledge-based question answering, intelligent search, educational question answering, biomedical knowledge exploration, and enterprise knowledge management. In these scenarios, answers often depend on multiple entities and relations rather than a single fact. By generating query-relevant subgraphs, SGR provides explicit evidence paths that help the language model produce more grounded and interpretable answers.
The framework is also useful for complex information retrieval. Compared with direct prompting or unstructured retrieval, schema-guided subgraph generation can preserve important relational structures while filtering out irrelevant information. This allows the model to focus on compact and useful evidence during reasoning.

However, SGR still produces several types of errors. First, entity linking errors may occur when the question contains ambiguous or rare entities. Second, relation identification errors may appear when natural language expressions do not match the relation labels in the knowledge graph. Third, useful triples may be removed during subgraph filtering, leading to incomplete reasoning paths. Finally, overly large subgraphs may introduce noisy evidence and mislead the model. These errors suggest that more accurate entity linking, relation matching, and path ranking strategies are needed in future work.

\section{Conclusion}

This paper proposed SGR, a stepwise reasoning enhancement framework that integrates LLMs with external knowledge graphs through query-relevant subgraph generation. By using schema-guided retrieval, Cypher-based reasoning, and collaborative answer integration, SGR grounds the reasoning process in structured evidence and improves interpretability. Experiments on CWQ, WebQSP, GrailQA, and KQA Pro show that SGR improves accuracy and Hits@1 over standard prompting and knowledge-enhanced baselines. Ablation results further confirm that schema guidance and Neo4j-based retrieval are key to the framework’s effectiveness. Overall, SGR demonstrates that external subgraphs can help LLMs perform more reliable and transparent multi-step reasoning.

\section*{Limitations}

SGR still has several limitations. First, its performance depends on the quality and coverage of the external knowledge graph; missing or noisy facts may lead to incorrect reasoning. Second, errors in entity extraction, relation identification, or schema construction can affect subgraph retrieval and final answers. Third, choosing an appropriate subgraph size is challenging, since small subgraphs may miss useful evidence while large ones may introduce noise. Finally, SGR requires additional computation for schema generation, Cypher querying, and reasoning-path integration, which may increase inference cost compared with direct prompting methods.

\bibliography{custom}

\end{document}